\documentclass{article}

\PassOptionsToPackage{numbers, compress}{natbib}

\usepackage[final]{neurips_2021}

\usepackage[utf8]{inputenc} 
\usepackage[T1]{fontenc}    
\usepackage{url}            
\usepackage{amsfonts}       
\usepackage{nicefrac}       
\usepackage{microtype}      
\usepackage{moreverb,url}
\usepackage[bookmarksopen,bookmarksnumbered,citecolor=black,urlcolor=black]{hyperref}
\usepackage{graphics} 
\usepackage{epsfig} 
\usepackage{times} 
\usepackage{amsmath} 
\usepackage{amssymb}  
\usepackage[ruled,vlined,linesnumbered]{algorithm2e}
\usepackage{float}
\usepackage{algpseudocode}
\usepackage{booktabs}
\usepackage{graphicx}
\usepackage[table,xcdraw,dvipsnames]{xcolor}
\usepackage{multirow}
\usepackage{xcolor,pifont}
\usepackage{flushend}
\usepackage{siunitx}

\let\oldnl\nl
\newcommand{\nonl}{\renewcommand{\nl}{\let\nl\oldnl}}

\usepackage{amsmath, amssymb}

\usepackage{xifthen}
\usepackage{color}

\newcommand{\ba}{\begin{eqnarray}}
\newcommand{\ea}{\end{eqnarray}}

\newcommand{\presup}[1]{\,{}^{\scriptscriptstyle #1}\!}

\newcommand{\pose}[1][ZZZZ]{\ifthenelse{\equal{#1}{ZZZZ}}{}{\presup{#1}}{\mathbf{\xi}}}
\newcommand{\estpose}[1][ZZZZ]{\ifthenelse{\equal{#1}{ZZZZ}}{}{\presup{#1}}{\mathbf{\hat{\xi}}}}
\newcommand{\hpose}[1][ZZZZ]{\ifthenelse{\equal{#1}{ZZZZ}}{}{\presup{#1}}{\hat{\mathbf{\xi}}}}
\newcommand{\posedot}[1][ZZZZ]{\ifthenelse{\equal{#1}{ZZZZ}}{}{\presup{#1}}{\mathbf{\nu}}}
\newcommand{\q}[1][ZZZZ]{\ifthenelse{\equal{#1}{ZZZZ}}{}{\presup{#1}}{\mathring{q}}}
\DeclareMathAlphabet{\mathitbf}{OML}{cmm}{b}{it}
\newcommand{\twist}[2][ZZZZ]{\ifthenelse{\equal{#1}{ZZZZ}}{}{\presup{#1}}{\mathcal{S}}}
\renewcommand{\vec}[2][ZZZZ]{\ifthenelse{\equal{#1}{ZZZZ}}{}{\presup{#1}}{\mathitbf{#2}}}
\newcommand{\hvec}[2][ZZZZ]{\ifthenelse{\equal{#1}{ZZZZ}}{}{\presup{#1}}{\tilde{\vec{#2}}}}
\newcommand{\evec}[2][ZZZZ]{\ifthenelse{\equal{#1}{ZZZZ}}{}{\presup{#1}}{\hat{\vec{#2}}}}
\newcommand{\bvec}[2][ZZZZ]{\ifthenelse{\equal{#1}{ZZZZ}}{}{\presup{#1}}{\bar{\vec{#2}}}}
\newcommand{\dvec}[2][ZZZZ]{\ifthenelse{\equal{#1}{ZZZZ}}{}{\presup{#1}}{\dot{\vec{#2}}}}
\newcommand{\ddvec}[2][ZZZZ]{\ifthenelse{\equal{#1}{ZZZZ}}{}{\presup{#1}}{\ddot{\vec{#2}}}}

\newcommand{\mat}[2][ZZZZ]{\ifthenelse{\equal{#1}{ZZZZ}}{}{\presup{#1}\,}{{\boldsymbol #2}}}
\newcommand{\dmat}[2][ZZZZ]{\ifthenelse{\equal{#1}{ZZZZ}}{}{\presup{#1}\,}{{\dot{\boldsymbol #2}}}}
\newcommand{\emat}[2][ZZZZ]{\ifthenelse{\equal{#1}{ZZZZ}}{}{\presup{#1}\,}{\hat{\boldsymbol#2}}}
\newcommand{\matfn}[3][ZZZZ]{\ifthenelse{\equal{#1}{ZZZZ}}{}{\presup{#1}}{{\mat{#2}}\left(#3\right)}}
\newcommand{\Rt}[2][ZZZZ]{\ifthenelse{\equal{#1}{ZZZZ}}{}{\presup{#1}}{{\bf R}\left(#2\right)}}

\newcommand{\point}[2][ZZZZ]{\ifthenelse{\equal{#1}{ZZZZ}}{}{\presup{#1}}{\mathbf{\mathrm{#2}}}}

\newfont{\School}{pncr}
\newfont{\eightTR}{pncr at 8pt}

\usepackage{color}
\usepackage{fancyvrb}
\fvset{formatcom=\color{blue},fontseries=c,fontfamily=courier,xleftmargin=4mm,commentchar=!}
\DefineVerbatimEnvironment{Code}{Verbatim}{formatcom=\color{blue},fontseries=c,fontfamily=courier,fontsize=\footnotesize,xleftmargin=4mm,commentchar=!}
\DefineVerbatimEnvironment{CodeSmall}{Verbatim}{formatcom=\color{blue},fontseries=c,fontfamily=courier,fontsize=\scriptsize,xleftmargin=1mm,commentchar=!}
\DefineVerbatimEnvironment{CodeNum}{Verbatim}{numbers=left,numbersep=4pt,formatcom=\color{blue},fontseries=c,fontfamily=courier,fontsize=\footnotesize,xleftmargin=4mm}

\newcommand{\model}[1]{\index{code}{#1@\textit{#1}}\ifthenelse{\boolean{draft}}{{\color{green}\Verb+#1+}}{\Verb+#1+}}
\newcommand{\block}[1]{\ifthenelse{\boolean{draft}}{{\color{green}\Verb+#1+}}{\textsf{#1}}}
\newcommand{\func}[2][ZZZZ]{\ifthenelse{\equal{#1}{ZZZZ}}{\index{code}{#2}}{\index{code}{#1}}\ifthenelse{\boolean{draft}}{{\color{green}\Verb+#2+}}{\Verb+#2+}}
\newcommand{\methodb}[2]{\index{code}{#1@\textbf{#1}!.#2}\ifthenelse{\boolean{draft}}{{\color{magenta}\Verb+#1.#2+}}{\Verb+#1.#2+}}
\newcommand{\method}[2]{\index{code}{#1@\textbf{#1}!.#2}\ifthenelse{\boolean{draft}}{{\color{magenta}\Verb+#2+}}{\Verb+#2+}}
\newcommand{\class}[1]{\index{code}{#1@\textbf{#1}}\ifthenelse{\boolean{draft}}{{\color{cyan}\Verb+#1+}}{\Verb+#1+}}
\newcommand{\property}[1]{\index{property}{#1}\ifthenelse{\boolean{draft}}{{\color{cyan}\Verb+#1+}}{\Verb+#1+}}

\newcommand\BibTeX{{\rmfamily B\kern-.05em \textsc{i\kern-.025em b}\kern-.08em
T\kern-.1667em\lower.7ex\hbox{E}\kern-.125emX}}

\title{Zero-Shot Uncertainty-Aware Deployment of Simulation Trained Policies on Real-World Robots}

\author{%
  Krishan Rana \\
  QUT Centre for Robotics, Brisbane\\
  \texttt{krishan.rana@hdr.qut.edu}
  \And
  Vibhavari Dasagi\\
  QUT Centre for Robotics, Brisbane\\
  \texttt{v.dasagi@hdr.qut.edu.au}
  \And
  Jesse Haviland\\
  QUT Centre for Robotics, Brisbane\\
  \texttt{j.haviland@hdr.qut.edu.au}
  \And
  Ben Talbot\\
  QUT Centre for Robotics, Brisbane\\
  \texttt{b.talbot@hdr.qut.edu.au}
  \And
  Michael Milford\\
  QUT Centre for Robotics, Brisbane\\
  \texttt{michael.milford@qut.edu.au}
  \And
  Niko S\"underhauf\\
  QUT Centre for Robotics, Brisbane\\
  \texttt{niko.suenderhauf@hdr.qut.edu.au}

}

\begin{document}

\maketitle

\begin{abstract}

While deep reinforcement learning (RL) agents have demonstrated incredible potential in attaining dexterous behaviours for robotics, they tend to make errors when deployed in the real world due to mismatches between the training and execution environments. In contrast, the classical robotics community have developed a range of controllers that can safely operate across most states in the real world given their explicit derivation. These controllers however lack the dexterity required for complex tasks given limitations in analytical modelling and approximations. In this paper, we propose Bayesian Controller Fusion (BCF), a novel uncertainty-aware deployment strategy that combines the strengths of deep RL policies and traditional handcrafted controllers. In this framework, we can perform zero-shot sim-to-real transfer, where our uncertainty based formulation allows the robot to reliably act within out-of-distribution states by leveraging the handcrafted controller while gaining the dexterity of the learned system otherwise. We show promising results on two real-world continuous control tasks, where BCF outperforms both the standalone policy and controller, surpassing what either can achieve independently. A supplementary video demonstrating our system is provided at \url{https://bit.ly/bcf_deploy}.
  
\end{abstract}

\section{Introduction}

As the adoption of autonomous robotic systems increases around us, there is a need for the controllers driving them to exhibit the level of sophistication required to operate in our everyday unstructured environments. Recent advances in reinforcement learning (RL) coupled with deep neural networks as function approximators, have shown impressive results across a range of complex robotic control tasks including dexterous in-hand manipulation \cite{dexterous}, quadrupedal locomotion \cite{haarnoja2018learning}, and targeted throwing \cite{throwing}. Nevertheless, the lack of \textit{safety} guarantees in deep RL-based controllers limits their usability for real-world robotics \cite{Ibarz_2021}. This comes as a result of the black-box nature of neural network policies and their inability to reliably deal with out-of-distribution states, particularly seen when simulation-trained models are transferred to the real world.

On the contrary, the classical robotics community have yielded numerous controllers and algorithms for a range of real-world physical systems (from mobile robots to humanoids) that allow us to reliably and safely deploy robotic agents in the real world. These include classical feedback controllers \cite{maxwell1868governors}, trajectory generators \cite{ijspeert2008central} and behaviour trees \cite{colledanchise2016behavior}. This is attributed to their explicit analytic derivation and known models, allowing for control theoretic guarantees which make them suitable for real-world deployment. They however can be highly suboptimal when applied to complex tasks, due to limitations in analytical modelling and approximations. 

A promising direction for the future of robot control is in combining the complementary strengths of these different control mechanisms in order to address their respective limitations. Such approaches have been observed in neuroscience, as the underlying control strategy used by biological systems. The dual-process theory of decision making \cite{dickinson2002role} proposes that multiple different neural controllers are involved when controlling action selection in biological systems. \citet{lee2014neural} provide evidence for this theory based on the human brain and show the existence of an arbitration mechanism that determines the extent to which the different neural controllers govern behaviours. The arbitrator bases its selection on specific performance measures exhibited by each controller, exploiting their respective strengths in a given state.

\begin{figure}[t]
  \centering
  \includegraphics[width=0.9\textwidth]{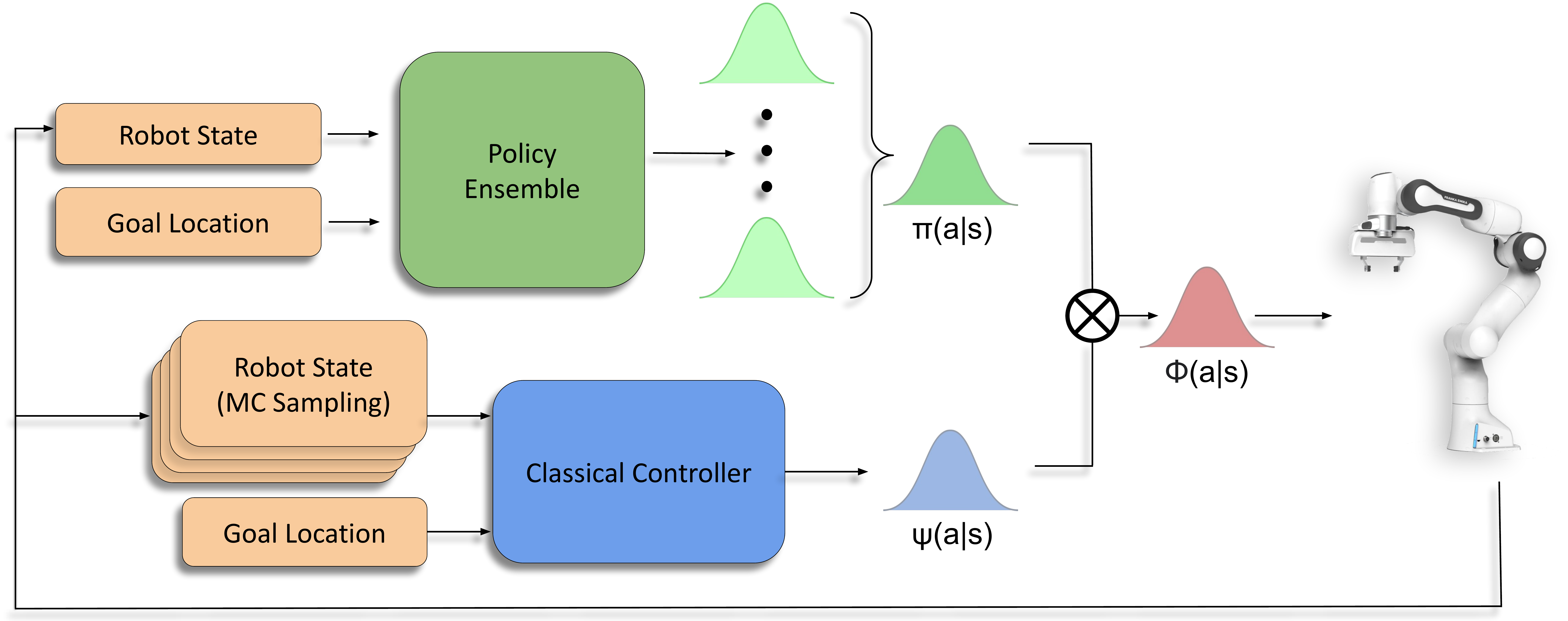}
  \caption{Bayesian Controller Fusion (BCF): a hybrid control strategy for safe deployment on real robotic systems. We derive uncertainty-aware action outputs for each controller and compose these outputs to better inform the action selection process.}
  \label{front}
\end{figure}

We draw inspiration from this observation and present Bayesian Controller Fusion (BCF), a hybrid control strategy that combines the respective strengths of deep RL and traditional handcrafted controllers (\textit{control priors}) for safe real-world deployment. We formulate the final policy as a Bayesian composition of these two controllers, where the controller output for each system captures their respective \textit{epistemic} uncertainty in a given state as shown in Figure \ref{front}. This allows BCF to naturally arbitrate control between the two systems based on their confidence to act. This has important implications during real-world deployment, where we gain the dexterity of the learned system in states that it has generalised to while relying on the risk-averse behaviours of the handcrafted system in out-of-distribution states for safe operation. Importantly, our method learns to control a real robot in joint space to complete a given task with no on-robot time (\textit{zero-shot}), even though the learned policy may not be perfectly transferable from simulation to the real world.

We demonstrate our approach on two continuous control, robotic tasks involving reactive navigation on a mobile robot and a manipulability maximising reacher on a robotic arm. We show how BCF allows us to reliably transfer a simulation trained policy to the real world while gaining significant performance improvements from the RL component when compared to the handcrafted system alone. We see this as a promising and practical avenue to bringing simulation-trained RL controllers to safely operate in the real world.

\section{Related Work}

Safe sim-to-real transfer has been an active research area, particularly in robotics where the cost of training robots directly in the real world is high. Many prior works have focused on developing realistic simulation environments that represent the real world as close as possible \cite{zhang2015towards, rusu2017sim}  or directly build a training environment from real-world data \cite{jakecorl}. While such approaches attempt to provide realistic environments, there are still a wide range of states and variations that are not captured by the training environment. Several works have generated robust policies using domain randomisation, where the agent is exposed to a wide range of environmental variations allowing it to generalise to changing environments. In the case of physical interaction with the environment, dynamics randomisation of simulated robots has also been used to capture the intricacies of real-world robots and their environments \cite{peng2018sim}. Recent works have also utilised a meta-RL approach where the agent's dynamics are adapted online in the real world \cite{arndt2020meta, exinet}. While all these approaches do produce increasingly robust policies for real-world operation, they still fail to capture the vast range of potential states and physical intricacies that the agent may encounter in the real world, limiting their safe operational space. 

As opposed to attempting to replicate the real world within the simulation environment, recent works have explored the ability of the agent to reason about the current state and utilise this as a proxy for decision making. \citet{osband2018randomized} and \citet{gal2016dropout} first explored the idea of state uncertainty estimation from neural networks to assist exploration during training. \citet{kahn2017uncertainty1} extended these ideas to the robotics setting to enable the agent to predict its state uncertainty as it acts within the real world. This allowed it to move slowly to avoid high-speed collisions while increasing its velocity within parts of the space where it had greater confidence in its actions. In our work, we explore an approach to avoid collisions altogether by switching to a safer controller. Other works learn a predictive model for catastrophic states using supervised learning \cite{lipton2016combating}, as well as a "backup policy" to return the agent from a critical state to a safe regime \cite{hans2008safe}. The operation of the robot is however constrained to the states that the predictor was trained on. As opposed to predictive models, we rely on epistemic uncertainty estimates directly from an ensemble of trained policies to identify unknown states. This removes the need to learn from labelled states or the restriction of operation within specific domains. More closely related to our work, \citet{garcia2012safe} utilise a distance-based risk estimator to identify out-of-distribution states and utilise this to switch between the learned policy and safe baseline controller. The switching however is abrupt and can result in jittery behaviours unsuitable for controlling real robots. In contrast, we formulate our controller as a composition of stochastic controllers allowing the resulting hybrid controller to smoothly interpolate between behaviours.

A growing area of interest is the combination of learning with traditional control strategies given their inherent safety guarantees. \citet{bansal20acombining} decouple the control from the perception module and learn an obstacle-free way-point finder that can be used by a low-level optimal controller for navigation. They show that this works well in the sim-to-real setting given the decomposition, however, the system is heavily reliant on the accuracy of the way-point prediction model. In our work we consider the uncertainty of our trained model to better inform the decision process. \citet{rana2020residual} leverage the Residual RL framework \cite{johannink2018residual, silver2018residual} to learn a dexterous reactive navigation controller and utilise uncertainty estimates of the residual policy to govern whether its action output is used to augment the behaviours of the underlying controller in the real world or not. The abrupt switching behaviour however resulted in noisy control signals not ideal for continuous control tasks. \citet{ryan_scaling2020} learn a range of skill competencies for a task and use a model predictive control (MPC) approach to forward simulate each of these skills in the simulation environment before executing the best action in the real world. While such hybrid controllers enable continuous and safe operation of learned policies in the real world, they come with a considerable computation overhead that can be detrimental to real-time operation. Our Bayesian fusion formulation allows us to directly leverage uncertainty estimates to govern the best action for execution between controllers at a given state, allowing for reactive and real-time control.

\section{Approach}

We introduce Bayesian Controller Fusion (BCF), a hybrid control strategy that composes stochastic action outputs from two separate control mechanisms: an RL policy $\pi(a|s)$, and a control prior $\psi(a|s)$. These outputs are formulated as distributions over actions, where each distribution captures the relative state uncertainty for the system to act. The Bayesian composition of these two outputs forms our hybrid policy $\phi(a|s)$. 

Figure \ref{front} illustrates our hybrid control strategy that composes the outputs from the learned policy $\pi(a|s)$ and control prior $\psi(a|s)$. The uncertainty-aware compositional policy $\phi(a|s)$ allows for the safe deployment of learned controllers. In states of high uncertainty, the compositional distribution naturally biases towards the reliable, risk-averse and potentially suboptimal behaviours suggested by the control prior. In states of lower uncertainty, it biases towards the policy, allowing the agent to exploit the optimal behaviours discovered by it. This is reminiscent of the arbitration mechanism suggested by \cite{lee2014neural} for behavioural control in the human brain, where the most confident controller assumes control in a given situation. This dual-control perspective provides a reliable strategy for bringing RL to real-world robotics, where generalisation to all states is near impossible and the presence of a risk-averse control prior serves as a reliable fallback.

\subsection{Method}
\label{method}
Given a policy, $\pi$ and control prior, $\psi$, we can obtain two independent estimates of an executable action, $a$, in a given state.
In a Bayesian context, we can utilise the normalised product to fuse these estimates under the assumption of a uniform prior, $p(a)$:
\begin{align}
    \label{eq_1}
    p\left(a \mid s, \mathbf{\theta}_{\pi}, \mathbf{\theta}_{\psi}\right)=\frac{p\left(s,\mathbf{\theta}_{\pi}, \mathbf{\theta}_{\psi} \mid a\right) p(a)}{p\left(s,\mathbf{\theta}_{\pi}, \mathbf{\theta}_{\psi}\right)}.
\end{align}

\noindent We assume Gaussian distributional outputs from each system and represent $\pi(a|s) \approx p\left(a\mid s, \theta_{\pi}\right)$ and $\psi(a|s) \approx p\left(a\mid s, \theta_\psi\right)$, where, $\theta_{\pi} = \{[\mu_{\pi_1},...,\mu_{\pi_n}]^\intercal, [\sigma_{\pi_1},...,\sigma_{\pi_n}]^\intercal\}$ and $\theta_{\psi} = \{[\mu_{\psi_1},...,\mu_{\psi_n}]^\intercal, [\sigma_{\psi_1},...,\sigma_{\psi_n}]^\intercal\}$ denote the distribution parameters for the policy and control prior outputs respectively and $n$ is the dimensionality of the action space. We drop the state $s$, to simplify notation.\\

\noindent Assuming statistical independence of $p\left(\mathbf{\theta}_{\pi} \mid a\right)$  and $p\left(\mathbf{\theta}_{\psi} \mid a\right)$, we can expand our likelihood estimate, $p\left(\mathbf{\theta}_{\pi}, \mathbf{\theta}_{\psi} \mid a\right)$, as follows:
\begin{align}
\label{eq_2}
    p\left(\mathbf{\theta}_{\pi}, \mathbf{\theta}_{\psi} \mid a\right)&=p\left(\mathbf{\theta}_{\pi} \mid a\right) p\left(\mathbf{\theta}_{\psi} \mid a\right)\nonumber \\
    &=\frac{p\left(a \mid \mathbf{\theta}_{\pi}\right) p\left(\mathbf{\theta}_{\pi}\right)}{p(a)} \frac{p\left(a \mid \mathbf{\theta}_{\psi}\right) p\left(\mathbf{\theta}_{\psi}\right)}{p(a)}.
\end{align}

\noindent Substituting this result back into (\ref{eq_1}), we can simplify the fusion as a normalised product of the respective action distributions from each control mechanism:

\begin{align}
  p\left(a \mid \mathbf{\theta}_{\pi}, \mathbf{\theta}_{\psi}\right)=\eta \underbrace{p\left(a \mid \mathbf{\theta}_{\pi}\right)}_{\substack{\text{Policy}}} \underbrace{p\left(a \mid \mathbf{\theta}_{\psi}\right)}_{\substack{\text{Control} \\ \text{Prior}}},
\end{align}
where,
\begin{align}
  \eta=\frac{p\left(\mathbf{\theta}_{\pi}\right) p\left(\mathbf{\theta}_{\psi}\right)}{p\left(\mathbf{\theta}_{\pi}, \mathbf{\theta}_{\psi}\right) p(a).} 
\end{align}



\noindent The composite distribution $p\left(a \mid \mathbf{\theta}_{\pi}, \mathbf{\theta}_{\psi}\right)$ forms our hybrid policy output\ $\phi(a|s)$. As we approximate the distributional output from each system to be univariate Gaussian for each action, the composite distribution $\phi(a|s)$ will also be univariate Gaussian $\phi(a|s) \sim \mathcal{N}(\mu_\phi, \sigma^{2}_\phi)$. As a result, we can compute the corresponding mean $\mu_{\phi}$ and variance $\sigma^{2}_{\phi}$  for each action as follows:
\begin{equation}
    \label{h_mu}
   \mu_\phi = \frac{\mu_{\pi}\sigma_{\psi}^{2} + \mu_{\psi}\sigma_{\pi}^{2}}{\sigma_{\psi}^{2} + \sigma_{\pi}^{2}} ,
\end{equation}

\begin{equation}
    \label{h_sig}
    \sigma_\phi^{2} = \frac{\sigma^{2}_{\pi}\sigma_{\psi}^{2}}{\sigma_{\psi}^{2} + \sigma_{\pi}^{2}} , 
\end{equation}

\noindent where this expansion implicitly handles the normalisation constant $\eta$. 

\subsection{Components}
\label{sec:comp}
In order to leverage our proposed approach in practice, we describe the derivation of the distributional action outputs for each system below and provide the complete BCF algorithm for combining these systems in Algorithm \ref{algorithm1}.

\subsubsection{Uncertainty-Aware Policy}
 We leverage stochastic RL algorithms that output each action as an independent Gaussian $\pi'(a|s)\sim \mathcal{N}(\mu_{\pi'}, \sigma_{\pi'}^2)$ where $\mu_{\pi'}$ denotes the mean and $\sigma^{2}_{\pi'}$ denotes the corresponding variance. This distribution is optimised to reflect the action which would both maximise the returns from a given state, as well as the entropy  \cite{haarnoja2019soft_}. Such exploration distributions tend to be risk seeking and do not capture the state uncertainty of the agent. The latter is a key component required for our BCF formulation. To attain an uncertainty-aware distribution, we leverage \textit{epistemic} uncertainty estimation techniques suggested in the computer vision literature based on ensemble learning \cite{lakshminarayanan2017simple}. We train an ensemble of $M$ agents to form a uniformly weighted Gaussian mixture model, and combine these predictions into a single univariate Gaussian whose mean and variance are respectively the mean, $\mu_{\pi}(s)$  and variance, $\sigma^{2}_{\pi}(s)$ of the mixture, $p(a | s, \theta_{\pi})=M^{-1} \sum_{m=1}^{M} p\left(a | s, \theta_{\pi'_m}\right)$ as described in \cite{lakshminarayanan2017simple}. The mean and variance of the mixture $M^{-1} \sum \mathcal{N}\left(\mu_{\pi'_m}(s), \sigma_{\pi'_m}^{2}(s)\right)$ are given by:
 
 \begin{equation}
 \label{fuse_mean}
\mu_{\pi}(s)=M^{-1} \sum_{m} \mu_{\pi'_{m}}(s)
\end{equation}

\begin{equation}
\label{fuse_sigma}
\sigma_{\pi}^{2}(s)=M^{-1} \sum_{m}\left(\sigma_{\pi'_{m}}^{2}(s)+\mu_{\pi'_{m}}^{2}(s)\right)-\mu_{\pi}^{2}(s)
\end{equation}
 
\noindent The empirical variance, $\sigma_{\pi}^{2}(s)$, of the resulting output distribution, $p(a | s,\theta_{\pi})$ approximates a measure of the policy's epistemic uncertainty in a given state for a particular action. This allows for a broader distribution when presented with unknown states and a tighter distribution in familiar states. This plays an important role within our BCF formulation as described previously. Note that we fuse the distributional polices as opposed to just their means to prevent collapse to a deterministic system once they converge. This allows the agent to continue to explore alternate actions and identify better solutions.

\subsubsection{Control Prior}

 In order to incorporate the inherently deterministic control priors developed by the robotics community within our stochastic RL framework, we require a distributional action output that captures its uncertainty to act in a given state.  As the uncertainty is state-centric, we empirically derive this action distribution by propagating noise (provided by the known sensor model variance, $\sigma^{2}_{\text{model}}$) from the sensor measurements through to the action outputs using Monte Carlo (MC) sampling. By computing the mean, $\mu_{\psi}$ and variance, $\sigma_{\psi}$ of the outputs, the distributional action output, $\mathcal{N}(\mu_{\psi}, \sigma_{\psi}^{2})$ for a given state, $s$ is given by:
 
\begin{equation}
    \label{cp_mu}
    \mu_{\psi}(s)=N^{-1} \sum_{n} a_{\psi}(s_{MC_{n}}), \hspace{20pt} s_{MC} \sim \mathcal{N}(s, \sigma_{\text{model}}^{2}),
\end{equation}

\begin{equation}
    \sigma_{\psi}^2(s) = N^{-1} \sum_{n} (a_{\psi}(s_{MC_{n}}) - \mu_{\psi}(s))^2,
\end{equation}
 
\noindent where $a_{\psi}(\cdot)$ denotes a deterministic action output from the control prior for a given MC sample and $N$ is the number of sampled states. Given the inherent robustness to noise of most control priors, we additionally set a minimum possible standard deviation for the distribution. This prevents the control prior distribution from collapsing to a deterministic value, rendering the policy useless within the BCF formulation. The resulting variance for the control prior distribution is defined as:
 
 \begin{align}
 \label{cp_distr}
    \sigma_{\psi}^2(s) =  \max\left( N^{-1} \sum_{n} (a_{\psi}(s_{MC_{n}}) - \mu_{\psi}(s))^2, \sigma_{d}^{2}(s)\right).
\end{align}
 
 \noindent The choice of $\sigma^2_d$ is left as a hyper-parameter for the user to set based on the specific controller used and its optimality towards solving the task.

\begin{algorithm}[t]
\SetAlgoLined
\textbf{Given:} Ensemble of \textit{M} policies ($[\pi'_{1}, \pi'_{2} ... \pi'_{M}]$), control prior ($\pi_{\psi}$) and variance ($\sigma_d^{2}$) \\
\KwIn{State $\textit{s}_t$}
\KwOut{Action $\textit{a}_t$}
  Approximate the policy ensemble predictions as a unimodal Gaussian $\pi(\cdot|s_{t}) \sim \mathcal{N}(\mu_{\pi}, \sigma_{\pi}^{2})$ described in Equations (\ref{fuse_mean}) and (\ref{fuse_sigma})\\
  
  Compute the control prior action distribution $\psi(\cdot|s_{t}) \sim \mathcal{N}\left(\mu_{\psi}, \sigma_{\psi}^{2}\right)$ as given in Equations (\ref{cp_distr}) and (\ref{cp_mu}) \\
  
  Compute the composite distribution $\phi(\cdot|s_{t}) \sim \mathcal{N}(\mu_\phi, \sigma^{2}_\phi)$\\
  
  \nonl \hspace{0.1cm} $\phi(\cdot|s_{t}) = \eta(\pi(\cdot|s_{t})\cdot\psi(\cdot|s_{t}))$ as given in Equations (\ref{h_mu}) and (\ref{h_sig})\\
  
  Select action $\textit{a}_t$ from the distribution $\phi(\cdot|s_{t})$\\
  
 \Return{$\textit{a}_t$}
 \caption{Bayesian Controller Fusion}
\label{algorithm1}
\end{algorithm}

Given the formulation for the distributional outputs from each system, we present the complete BCF algorithm for governing action selection both during training and deployment in Algorithm \ref{algorithm1}.

\begin{figure}[t]
  \centering
  \includegraphics[width=0.99\textwidth]{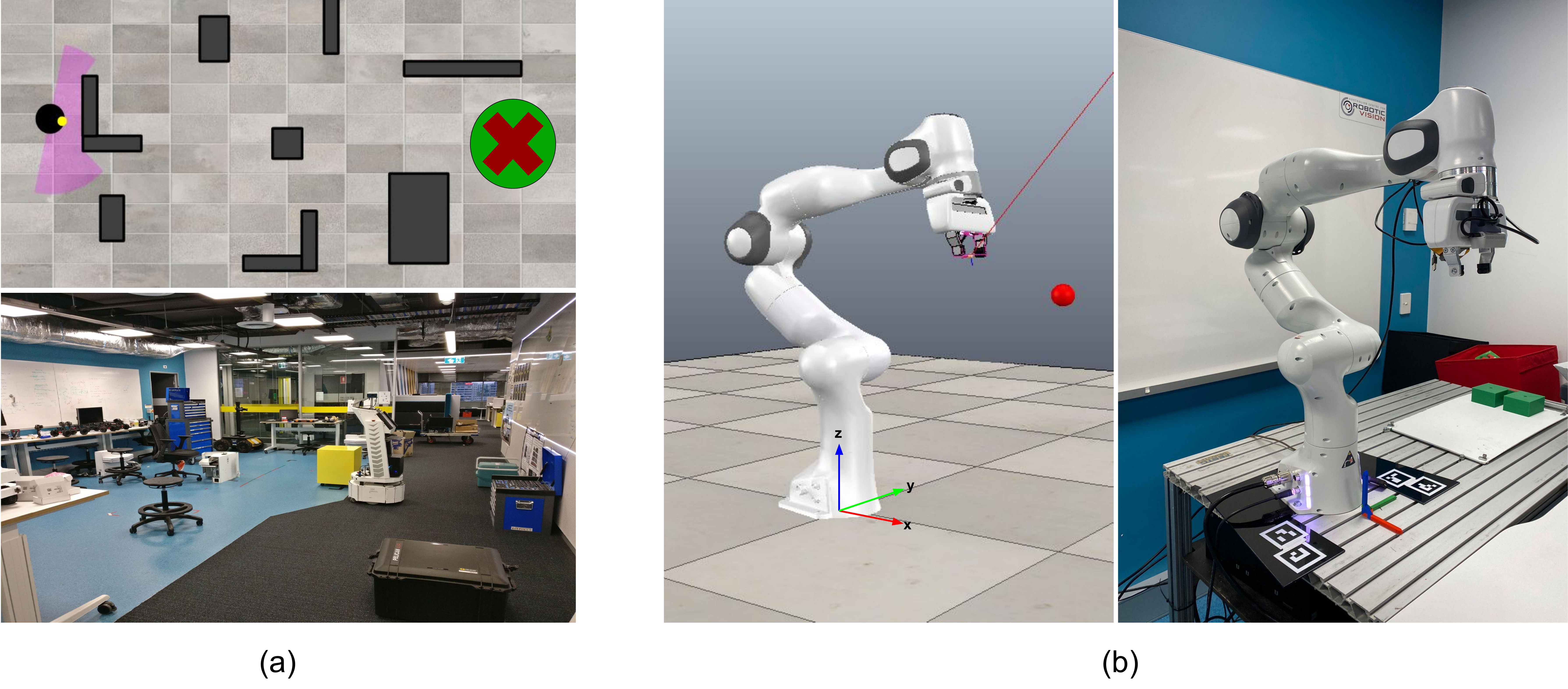}
  \caption{Simulation training environments and real world deployment environments for (a) PointGoal Navigation and (b) Maximum Manipulability Reacher tasks. Note the stark discrepancy in obstacle profiles for the navigation task between the simulation environment and real world environments.}
  \label{robots}
\end{figure}

\section{Experiments}

\label{sec:evaluation_deployed}
In this section, we assess the ability of BCF to reliably control real-world robots with a simulation trained policy and without any on-robot fine-tuning. We evaluate our approach on two continuous control robotics tasks: target driven navigation, and a manipulability maximising reacher task. We provide a detailed description of these tasks in Appendix \ref{tasks}. We additionally compare BCF to its individual learned and handcrafted components in isolation in order to understand its ability to exploit their respective strengths. We provide details of the evaluated systems below.

\begin{enumerate} 
    \item \textit{Control Prior:} The deterministic classical controller derived using analytic methods.
    \item \textit{SAC:} RL policy trained using vanilla Soft Actor Critic (SAC) \cite{haarnoja2018soft}.
    \item \textit{BCF:} Our proposed hybrid control strategy that combines uncertainty aware outputs from the control prior and the learned RL policy. The agent was trained using SAC and Algorithm \ref{algorithm1} for action selection during exploration.
\end{enumerate}

Note that all the policies used in this evaluation were trained to convergence in the simulation environments. We provide a detailed account for each task in the following sections.

\subsection{PointGoal Navigation} In this experiment, we examine whether BCF could overcome the limitations of an existing reactive navigation controller, in this case, an Artificial Potential Fields (APF) controller \cite{warren1989global, koren1991potential}, while leveraging this control prior to safely deal with out-of-distribution states that the policy could fail in. The APF controller used exhibited suboptimal oscillatory behaviours particularly in between obstacles.

\begin{table}[t]
\centering
\caption{Evaluation of PointGoal Navigation in the Real World}
\label{tab:real_robot}
\resizebox{0.94\textwidth}{!}{%
\begin{tabular}{@{}lcccc@{}}
\toprule
              & \multicolumn{2}{c}{\textbf{Trajectory 1}} & \multicolumn{2}{c}{\textbf{Trajectory 2}} \\ \midrule
\multicolumn{1}{c}{\textbf{Method}} &
  \textbf{\begin{tabular}[c]{@{}c@{}}Distance Travelled \\ (meters)\end{tabular}} &
  \textbf{\begin{tabular}[c]{@{}c@{}}Actuation Time \\ (seconds)\end{tabular}} &
  \textbf{\begin{tabular}[c]{@{}c@{}}Distance Travelled \\ (meters)\end{tabular}} &
  \textbf{\begin{tabular}[c]{@{}c@{}}Actuation Time \\ (seconds)\end{tabular}} \\ \midrule
Control Prior & 42.3                & 274                 & 35.3                & 277                 \\
SAC   & Fail                & Fail                & Fail                & Fail                \\
Move-Base     & 62.6                & 263                 & 35.8                & 258                 \\
\textbf{BCF}  & \textbf{41.2}                & \textbf{135}                 & \textbf{30.4 }               & \textbf{117 }                \\ \bottomrule
\end{tabular}%
}
\end{table}

\begin{figure}[t]
  \centering
  \includegraphics[width=0.94\textwidth]{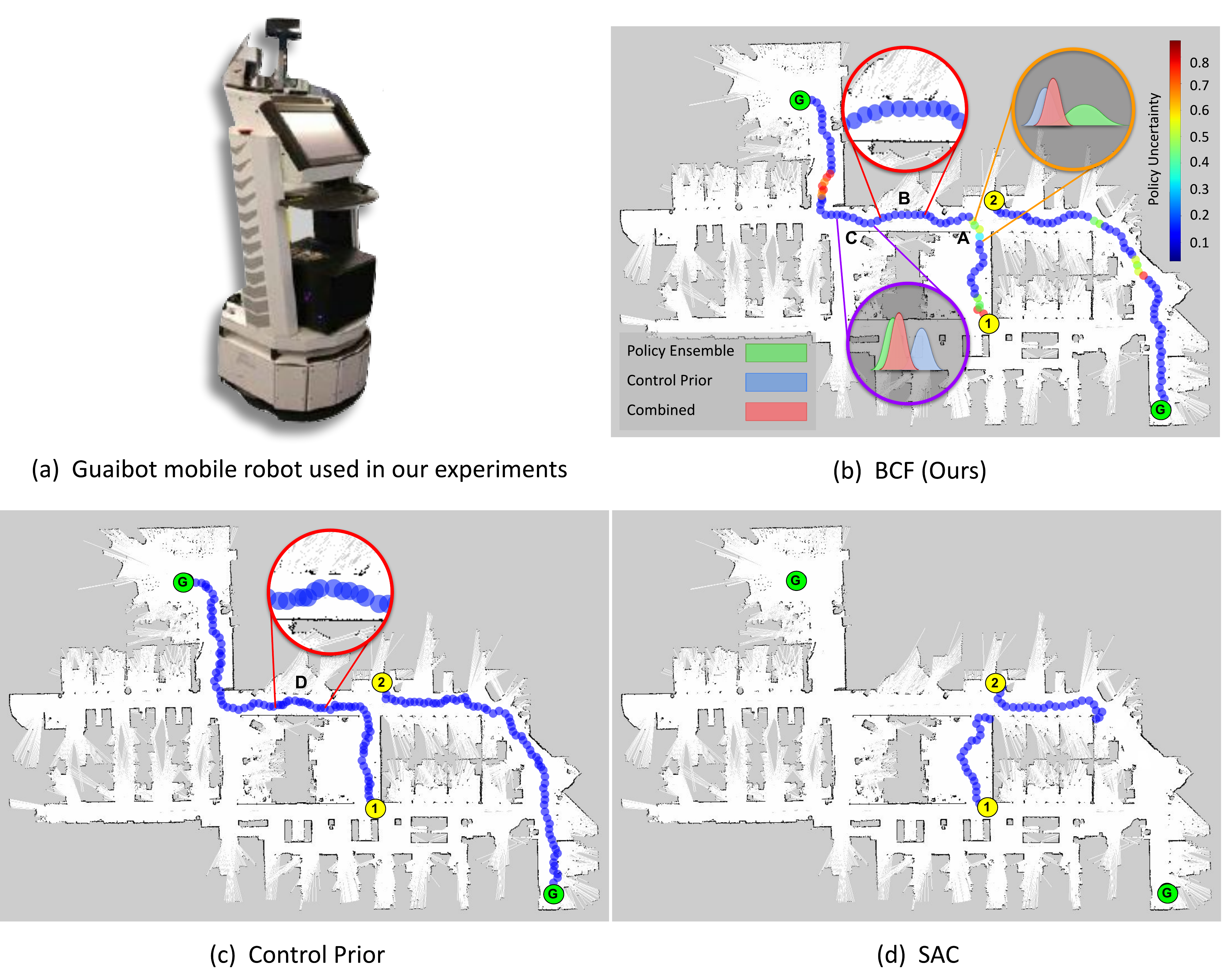}
  \caption{Trajectories taken by the real robot for different start and goal locations in a cluttered office environment with long narrow corridors. The trajectory was considered unsuccessful if a collision occurred. The trajectory taken by BCF is colour coded to represent the uncertainty in the linear velocity of the trained policy. We illustrate the behaviour of the fused distributions at key areas along the trajectory. The symbols \textbf{1} and \textbf{2} indicate the start locations for each trajectory and \textbf{G} indicates the corresponding goal locations.}
  \label{maps}
\end{figure}

\subsubsection{Evaluation}
\label{sec:real_world}

We utilise a GuiaBot mobile robot which is equipped with a 180$^\circ$ laser scanner, matching that used in the simulation environment. The velocity outputs from the policies are scaled to a maximum of \SI{0.25}{\meter/\second} before execution on the robot at a rate of 100 Hz. The system was deployed in a cluttered indoor office space that was previously mapped using the laser scanner. We utilise the ROS \textit{AMCL} package to localise the robot within this map and extract the necessary state inputs for the policy network and control prior. Despite having a global map, the agent is only provided with global pose information with no additional information about its operational space. The environment also contained clutter which was unaccounted for in the mapping process. To enable large traversals through the office space, we utilise a global planner to generate target sub-goals, for our reactive agents to navigate towards. We report the distance travelled by each controller and compare them to the distance travelled by a fine-tuned ROS \textit{move-base} controller. This controller is not necessarily the optimal solution but serves as a practical example of a commonly used controller on the Guiabot.

The evaluation was conducted on two different trajectories indicated as Trajectory 1 and 2 in Figure \ref{maps} and Table \ref{tab:real_robot}. Trajectory 1 consisted of a lab space with multiple obstacles, tight turns, and dynamic human subjects along the trajectory, while Trajectory 2 consisted of narrow corridors never seen by the robot during training. We terminated a trajectory once a collision occurred and marked the run as a failed attempt. We summarise the results in Table \ref{tab:real_robot}. 

Across both trajectories, the standalone SAC agent failed to complete a trajectory without any collisions, exhibiting sporadic reversing behaviours in out-of-distribution states. We can attribute these behaviours to its poor generalisation in such states, given the discrepancies in obstacle profiles seen during training in simulation and those encountered in the real world as shown in Figure \ref{robots} (a). The control prior was capable of completing all trajectories however required excessive actuation times. We can attribute this to its inefficient oscillatory motion when moving through passageways and in between obstacles. BCF was successful across both trajectories exhibiting the lowest actuation times across all methods. This indicates its ability to exploit the optimal behaviours learned by the agent while ensuring it did not act sporadically when presented with out-of-distribution states. It also demonstrates superior results when compared with the fine-tuned ROS~\textit{move-base} controller.

To gain a better understanding of the reasons for BCF's success when compared to the control prior and SAC agent acting in isolation, we examine the trajectories taken by these systems as shown in Figure \ref{maps}. The trajectory attained using BCF is colour-coded to illustrate the uncertainty of the policy's actions as given by the outputs of the ensemble. We draw the readers attention to the region marked \textbf{A} which exhibits higher values of policy uncertainty. The composition of the respective distributions at this region is shown within the orange ring. Given the higher policy uncertainty at this point, the resulting composite distribution was biased more towards the control prior which displayed greater certainty, allowing the robot to progress beyond this point safely. We note here that this is the particular region that the SAC agent failed as shown in Figure \ref{maps} (c). The purple ring at region \textbf{C} illustrates a region of low policy uncertainty with the composite distribution biased closer towards the policy. Comparing the performance benefit over the control prior gained in such a case, we draw the readers attention to regions \textbf{B} and \textbf{D} which show the path profile taken by the respective agents. The dense darker path shown by the control prior indicates regions of high oscillatory behaviour and significant time spent at a given location. On the other hand, we see that BCF does not exhibit this and attains a smoother trajectory which is attributed to the learned policy having higher precedence in these regions, stabilising the oscillatory effects of the control prior. This illustrates the ability of BCF to exploit the relative strengths of each component throughout deployment.

\subsection{Maximum Manipulability Reacher}

We evaluate the ability of BCF to build upon the basic structure provided by a Resolved Rate Motion Controller (RRMC) \cite{rrmc}, for reaching on a 7 DoF arm robot, in order to learn a more complex manipulability maximising reaching controller. While RRMC provides the policy with the basic knowledge to reach a goal, the agent has to learn how to modify the individual velocities of each joint in order to maximise the manipulability of the controller. More details on the tasks are provided in Appendix \ref{task_rrmc}.

\begin{table}[!t]
\centering
\caption{Evaluation of Maximum Manipulability Reacher in the Real World}
\label{tab:my-table}
\resizebox{0.91\textwidth}{!}{%
\begin{tabular}{@{}lccc@{}}
\toprule
\textbf{Method} &
  \multicolumn{1}{l}{\textbf{Average Manipulability}} &
  \multicolumn{1}{l}{\textbf{Average Final Manipulability}} &
  \multicolumn{1}{l}{\textbf{Success Rate}} \\ \midrule
Control Prior & 0.0629 $\pm$ 0.00926        & 0.0658 $\pm$ 0.0165          & 98.2\%          \\
SAC           & 0.0803 $\pm$ 0.00514         & 0.07812 $\pm$ 0.0150         & 78.6\%          \\
\textbf{BCF}  & \textbf{0.0836 $\pm$ 0.0156} & \textbf{0.0889 $\pm$ 0.0177} & \textbf{98.2\%} \\ \bottomrule
\end{tabular}%
}
\end{table}

\begin{figure}[!ht]
  \centering
  \includegraphics[width=0.99\textwidth]{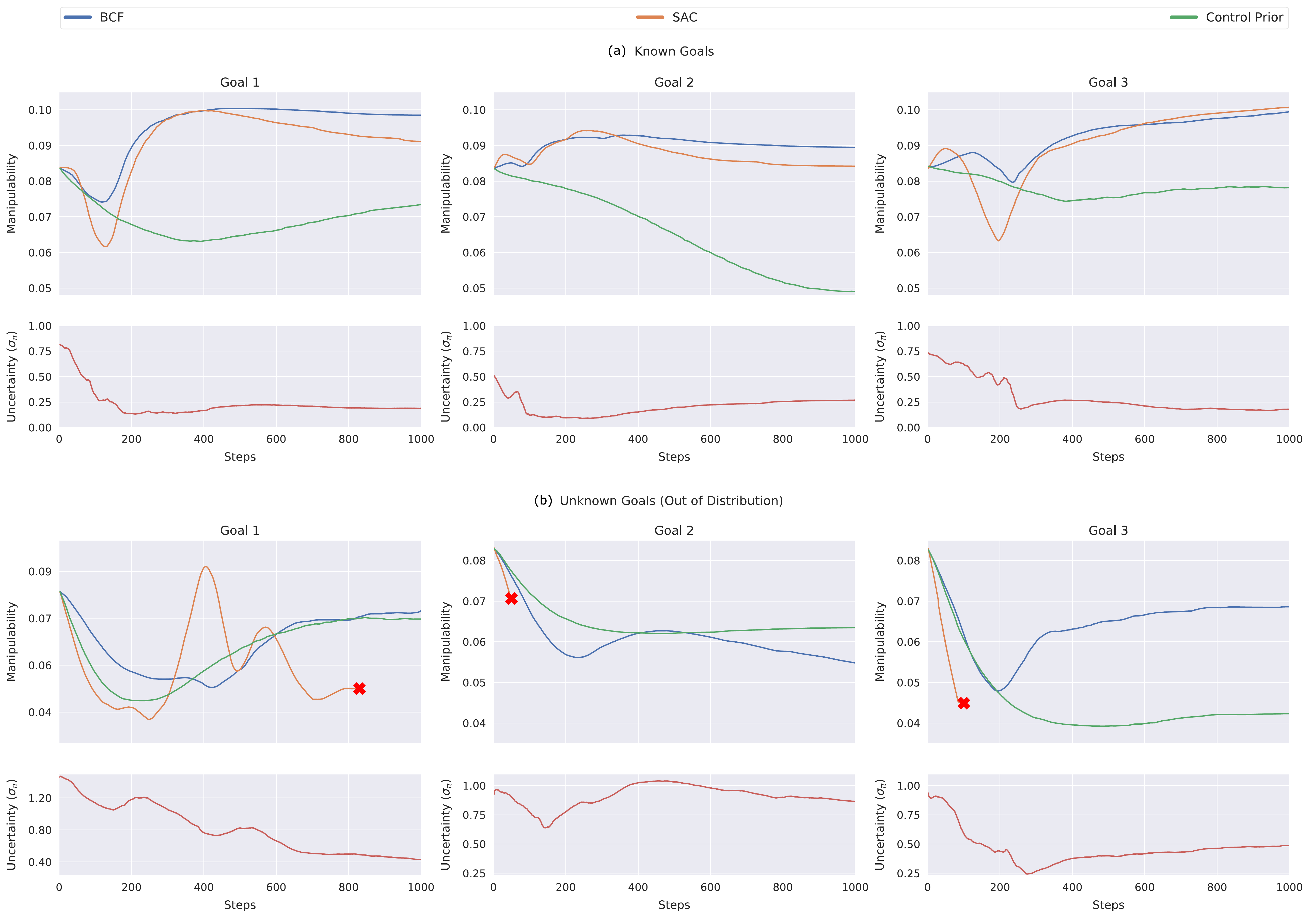}
  \caption{Manipulability and uncertainty curves for known and out-of-distribution goals for the reacher task, deployed on a real robot. The red cross indicates a failed trajectory.}
  \label{robot_manip_curves}
\end{figure}

\subsubsection{Evaluation}
To ensure that the simulation trained policies could be transferred directly to a real robot, we matched the coordinate frames of the PyRep simulator \cite{james2019pyrep} used with the real Franka Emika Panda robot setup shown in Figure \ref{robots}. The state and action space were matched with that used in the training environment, with the actions all scaled down to a maximum of \SI{1.74}{\radian\per\second} before publishing them to the robot at a rate of 100 Hz. The robot was trained with a subset of goals randomly sampled from the positive x-axis region of its workspace as shown in Figure \ref{robots}. We classify goal states sampled from outside this region as out-of-distribution states.

Table \ref{tab:my-table} shows the results obtained when evaluating the agent on a random set of ten different goals sampled from the robot's entire workspace. We report the average manipulability across the entire trajectory for all sampled goals as well as the average final manipulability attained at the end of all trajectories. We additionally indicate the success rate for each controller to reach the given goals. In all cases, BCF attains the highest manipulability and success rate surpassing both the control prior and SAC policy. It additionally illustrates its ability to deal with higher dimensional action spaces.

To better understand how BCF attains successful trajectories when compared to a standalone SAC policy, we take a closer look at the individual trajectories taken by the robot for goals sampled from both the known and out-of-distribution regions. From each region, we sampled 3 goals and show the corresponding manipulability curves of the robot across the trajectory in Figure \ref{robot_manip_curves}. For each goal, we additionally plot the policy ensemble uncertainty estimate used by BCF across the trajectory as indicated by the red curves. As shown in Figure \ref{robot_manip_curves} (a), for the known goals, BCF and the SAC agent both attain similar performances, maximising the manipulability of the agent across the trajectory. This is in stark contrast to the control prior which exhibits poor performance. Note here that while the control prior exhibits poor performance with regard to manipulability, it is still successful in completing the reaching task at hand without any failures. It is interesting to note the high uncertainty of the ensemble at the start of a trajectory which quickly drops to a significantly lower value. The high uncertainty is a result of the multiple possible trajectories that the robot could take at the start, which quickly narrows down once the robot begins to move. Note that once the policy ensemble exhibits a lower uncertainty, the corresponding performance of BCF closely resembles that of the standalone SAC agent, indicating that BCF does not cripple the optimality of the learned policy.

When evaluating the agents on out-of-distribution goals as shown in Figure \ref{robot_manip_curves} (b), BCF plays an important role in ensuring that the robot can successfully and safely complete the task. Note the higher levels of uncertainty across these trajectories when compared to the known goals case. In all these cases, the standalone SAC agent fails to successfully complete a trajectory, frequently self-colliding or exhibiting random sporadic behaviours. We indicate these failed trajectories with a red cross in Figure \ref{robot_manip_curves} (b). BCF is seen to closely follow the behaviours of the control prior in states of high uncertainty, averting it from such catastrophic failures. While the composite control strategy works well to ensure the safety of the robot, the higher reliance of the system on the control prior results in suboptimal behaviour with regards to manipulability. We provide a supplementary video to demonstrate these behaviours \footnote{Video demonstration: \url{https://bit.ly/bcf_deploy}}. The trade-off between task optimality versus the safety of the robot is an interesting dilemma that BCF attempts to balance naturally. The fixed variance, $\sigma_d^{2}$ chosen for the prior controller could serve as a tuning parameter to allow the user to control this trade-off at deployment. A smaller variance would bias the resulting controller more strongly towards the control prior yielding more conservative and suboptimal actions; whereas a larger variance would allow for close to optimal behaviours at the expense of the robots safety. We leave the exploration of this idea to future work.

\section{Discussion and Future Work}
\label{sec:conclusion}

Building on the large body of work already developed by the robotics community can greatly help accelerate the use of RL based systems, allowing us to develop better controllers for robots as they move towards solving tasks in the real world. The ideas presented in this paper demonstrate a strategy that closely couples traditional controllers with learned systems, exploiting the strengths of each approach in order to attain more reliable and robust behaviours in the zero-shot sim-to-real setting. We see this as a promising step towards safely bringing reinforcement learning to real-world robotics.

Across two robotics tasks for navigation and reaching, we show that BCF can safely deal with out-of-distribution states in the sim-to-real setting without any fine-tuning, succeeding where a typical standalone policy would fail, while attaining the optimality of the learned behaviours in known states. In the navigation domain, we overcome the inefficient oscillatory motion of an existing reactive navigation controller, decreasing the overall actuation time during real-world navigation runs by 50.7\%. For the reaching task, we show that our hybrid controller achieves the highest success rate, and improves the manipulability of an existing reaching controller by 34.9\%, a controller typically difficult to attain using analytical approaches.

While the uncertainty-based compositional policy we derive using BCF does train with the control prior in the loop, the policy is not directly aware of the control prior's presence. This could impact its overall ability to work in synergy with the control prior at deployment. In future work, we propose to incorporate the control prior in the Q-value update or alternatively learn a gating parameter to better inform the fusion process. This should allow the hybrid controller to operate on more complex tasks as well as interpolate across multiple behaviours. We are also interested in exploring alternative state uncertainty estimation techniques for both the control prior and RL that are faster than the sampling based approaches used in this work. This includes work from the supervised learning literature for out-of-distribution detection and distance-based uncertainty estimation techniques \cite{Miller_2021_WACV}.


\bibliographystyle{plainnat}
\bibliography{references.bib,manual.bib}

\appendix
\section{Appendix}

\subsection{Task Description}
\label{tasks}

\subsubsection{PointGoal Navigation}
The objective of this task is to navigate a robot from a start location to a goal location in the shortest time possible, while avoiding obstacles along the way. We utilise the training environment provided by \cite{rana2020residual}, which consists of five arenas with different configurations of obstacles. The goal and start location of the robot are randomised at the start of every episode, each placed on the extreme opposite ends of the arena (see Figure~\ref{robots} (a)). This sets the long horizon nature of the task. As we focus on the sparse reward setting, we define $r(s_t,a_t,s_{t+1}) = 1$ if $d_{\text{target}} < d_{\text{threshold}} $ and $r(s_t,a_t,s_{t+1}) = 0$ otherwise, where $d_{\text{target}}$ is the distance between the agent and the goal and $d_{\text{threshold}}$ is a set threshold. The action $a_t$ consists of two continuous values: linear velocity $\nu_{\text{nav}}\in[-1,1]$ and angular velocity $\omega_{\text{nav}}\in[-1,1]$. We assume that the robot can localise itself within a global map in order to determine its relative position to a goal location. The 180$^{\circ}$ laser scan range data is divided into 15 bins and concatenated to the robot's angle. The overall state $s_{t}$ of the environment is comprised of:
\begin{itemize}
    \item The binned laser scan data $l_{\text{bin}} \in \mathbb{R}^{15}$,
    \item The pose error between the robot's pose and the goal location $e_{t} \in \mathbb{R}^2$,
    \item The previous executed linear and angular velocity $a_{t-1} \in \mathbb{R}^2$,
\end{itemize}
for a total of 19 dimensions. The length of each episode is set to a maximum of 500 steps and does not terminate once the goal is achieved.

\subsubsection{Manipulability Maximising Reacher}
\label{task_rrmc}
The objective of this task is to actuate each joint of a manipulator within a closed-loop velocity controller such that the end-effector moves towards and reaches the goal point, while the manipulability of the manipulator is maximised. The manipulability index describes how easily the manipulator can achieve any arbitrary velocity. The ability of the manipulator to achieve an arbitrary end-effector velocity is a function of the manipulator Jacobian. While there are many methods that seek to summarise this, the manipulability index proposed by \cite{manip} is the most used and accepted within the robotics community \cite{manip2}. Utilising Jacobian based indices in existing controllers have several limitations, require greater engineering effort than simple inverse kinematics based reaching systems, and precise tuning in order to ensure the system is operational \cite{mmc}. We explore the use of RL to learn such behaviours by leveraging simple reaching controllers as priors. We utilise the PyRep simulation environment \cite{james2019pyrep}, with the Franka Emika Panda as our manipulator as shown in Figure \ref{robots} (b).  For this task, we generate a random initial joint configuration and random end-effector goal pose. We use a sparse goal reward, $r(s_t,a_t,s_{t+1}) = 1$ if $e_{t} < e_{\text{threshold}} $ and $r(s_t,a_t,s_{t+1}) = m$ otherwise, where $e_{t}$ is the spatial translational error between the end-effector and the goal, $e_{\text{threshold}}$ is a set threshold and
\begin{equation}
    m = \sqrt{ 
            \mbox{det}
            \left(
                J(q) J(q)^\top
            \right) 
    }  \ \in [0,\infty)
\end{equation}
is the manipulability of the robot at the particular joint configuration $q$ where $J(q)$ is the manipulator Jacobian. The action space consists of the manipulator joint velocities $\dot{\it{q}} \in [-1,1]^n$, where the values are continuous, and $n$ is the number of joints within the manipulator. In this work, the manipulator used consists of 7 joints. The state, $s_t$, of the environment is comprised of:
\begin{itemize}
    \item The joint coordinate vector $\it{q} \in \mathbb{R}^7$,
    \item The joint velocity vector $\dot{\it{q}} \in \mathbb{R}^7$,
    \item The translation error between the manipulator's end-effector and the goal $e_t(\it{q}) \in \mathbb{R}^3$,
    \item The end-effector translation vector $e_p(\it{q}) \in \mathbb{R}^3$,
\end{itemize}
for a total of 20 dimensions. Similar to the navigation task, the episode length for this task in fixed at 1000 steps and only terminates at the end of the episode.

\end{document}